\def\year{2021}\relax
\documentclass[letterpaper]{article} % DO NOT CHANGE THIS
\usepackage{aaai21}  % DO NOT CHANGE THIS
\usepackage{times}  % DO NOT CHANGE THIS
\usepackage{helvet} % DO NOT CHANGE THIS
\usepackage{courier}  % DO NOT CHANGE THIS
\usepackage[hyphens]{url}  % DO NOT CHANGE THIS
\usepackage{graphicx} % DO NOT CHANGE THIS
\usepackage{natbib}  % DO NOT CHANGE THIS AND DO NOT ADD ANY OPTIONS TO IT
\usepackage{caption} % DO NOT CHANGE THIS AND DO NOT ADD ANY OPTIONS TO IT
\usepackage{algorithm}
\usepackage{algorithmic}
\usepackage{subcaption}
\usepackage{amsmath}
\usepackage{booktabs} % For resolving the table errors[ 
\usepackage{bigfoot}

\title{Guiding Robot Exploration in Reinforcement Learning via Automated Planning}
\author{
    Yohei Hayamizu\textsuperscript{\rm 1}\thanks{Research conducted while visiting SUNY Binghamton.},
    Saeid Amiri\textsuperscript{\rm 2}, 
    Kishan Chandan\textsuperscript{\rm 2}, 
    Keiki Takadama\textsuperscript{\rm 1}, 
    Shiqi Zhang\textsuperscript{\rm 2}
    \\
}

\affiliations{
    \textsuperscript{\rm 1} The University of Electro-Communications, Tokyo, Japan\\
    \textit{hayamizu@cas.lab.uec.ac.jp, keiki@inf.uec.ac.jp} \\
    \textsuperscript{\rm 2} The State University of New York at Binghamton, Binghamton, NY, USA \\
    \textit{\{samiri1; kchanda2; zhangs\}@binghamton.edu}
}

\begin{document}

\maketitle

\begin{abstract}

Reinforcement learning (RL) enables an agent to learn from trial-and-error experiences toward achieving long-term goals; automated planning aims to compute plans for accomplishing tasks using action knowledge. 
Despite their shared goal of completing complex tasks, the development of RL and automated planning has been largely isolated due to their different computational modalities. 
Focusing on improving RL agents' learning efficiency, we develop Guided Dyna-Q (GDQ) to enable RL agents to reason with action knowledge to avoid exploring less-relevant states. 
The action knowledge is used for generating artificial experiences from an optimistic simulation. 
GDQ has been evaluated in simulation and using a mobile robot conducting navigation tasks in a multi-room office environment. 
Compared with competitive baselines, GDQ significantly reduces the effort in exploration while improving the quality of learned policies.

\end{abstract}

\section{Introduction}

Recent advances in artificial intelligence have enabled robots to conduct a variety of service and interaction tasks in human-inhabited environments~\cite{hawes2017strands,khandelwal2017bwibots,veloso2018increasingly}. 
When a world model of dynamics and rewards is available, one can use Markov decision process (MDP) algorithms to compute action policies~\cite{puterman2014markov}.
In practice, however, world models are frequently unavailable or tend to change over time due to exogenous changes. 
Reinforcement learning (RL) algorithms have been used to help agents learn action policies from trial-and-error experiences toward maximizing long-term utilities~\cite{sutton2018reinforcement}. 

There are various types of RL algorithms. 
Model-based RL enables agents to learn a world model while learning an action policy to achieve long-term goals~\cite{brafman2002r,mann2011scaling,Kaiser2020Model}. 
Model-based RL can easily incorporate domain knowledge, such as world dynamics, from a human expert into the process of policy learning. 
In addition, model-based RL is goal-independent from the model construction perspective, rendering the learned world model applicable to other tasks. 
The learned model cannot represent all dynamics, and thus, model-based RL can be susceptible to domain changes.
We are still interested in model-based RL in this work acknowledging its limitations, due to the characteristics of service robotics domains, such as widely available domain knowledge (e.g., how rooms are connected through doors), and highly diverse service requests (e.g., requests of guiding visitors from and to different indoor locations).

\begin{figure*}[t]
    \begin{center}
    \includegraphics[width=.7\linewidth]{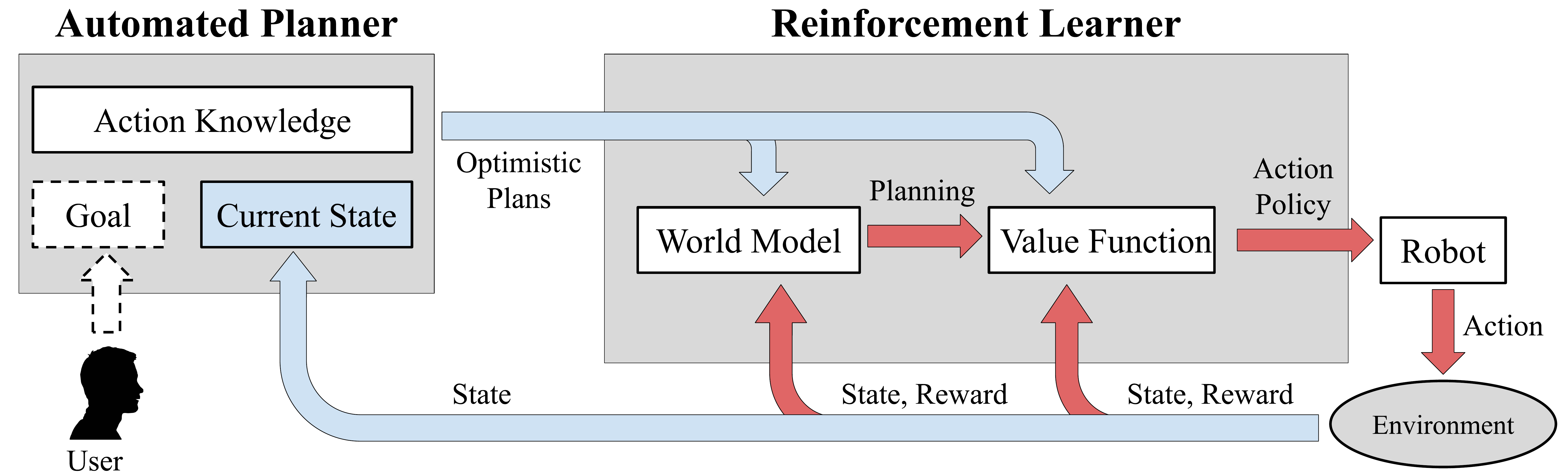}
    \end{center}
    \caption{An overview of Guided Dyna-Q (GDQ), where the key is the interplay between an automated planner and a reinforcement learner. The 
    red-color loop 
    corresponds to the standard control loop of Dyna-Q, where the agent (robot) interacts with the environment to update both its world model, and its $Q$-value function. Beyond that, GDQ further incorporates an automated planner into the learning process in 
    the 
    blue-color loop
    where goal-independent action knowledge (highly sparse, and potentially inaccurate)  is used for computing action sequences toward goal achievement. The action sequences are then used for updating both the world model and the $Q$-value function. }
    \label{fig:overview}
\end{figure*}

In this paper, we focus on addressing the low sample-efficiency challenge of model-based RL algorithms. 
We develop Guided Dyna-Q (\textbf{GDQ}) that consolidates the two classical paradigms of model-based RL and automated planning to help the agent avoid exploring less-relevant states toward more sample-efficient model learning and decision making. 
GDQ reasons with action knowledge to optimistically simulate action sequences to ``accomplish'' tasks. 
The simulated experience is then used to initialize and update Q-values toward efficient policy learning. 
In particular, we use Answer Set Programming (ASP) to formulate action knowledge~\cite{lifschitz2019answer}, and use Dyna-Q for model-based RL~\cite{sutton2018reinforcement}, though GDQ is not restricted to particular planning or learning paradigms. 
It should be noted that we only use widely available action knowledge, such as ``\emph{To open a door, one has to be in front of it first,}'' where knowledge acquisition is not a problem. 
The goal is to improve the learning efficiency in service robotics domains, and show that GDQ can leverage the complementary features of learning and planning paradigms to produce the best performance. 
We summarize GDQ, including the interplay between RL and automated planning, in Figure~\ref{fig:overview}. 

We have evaluated GDQ in simulation, and demonstrated the learning process using a real mobile robot. 
Results show that GDQ significantly improves the learning efficiency in comparison to existing model-based and model-free RL methods, including Q-Learning and Dyna-Q~\cite{sutton2018reinforcement}, as well as a competitive baseline that uses action knowledge to guide RL~\cite{DBLP:journals/ai/LeonettiIS16}. 
In a real-world office environment with more than 20 rooms, GDQ helped a mobile robot learn the optimal solution from only 30 episodes, whereas vanilla Dyna-Q could not find a meaningful solution in a reasonable runtime.

\section{Background}
\label{sec:back}

In this section, we briefly summarize the concepts of reinforcement learning and automated planning.

\subsection{Reinforcement Learning}
\label{sec:rl}

Within the MDP context, the agent must learn action policies from trial-and-error experiences when world models (reward functions, transition functions, or both) are not known. 
For instance, Q-learning is a model-free RL algorithm, and its $Q$-value function can be updated as below. 
% \begin{small}
\begin{equation*}
    \label{eq:updateQ}
    Q(s, a) \leftarrow Q(s, a) + \alpha \left[r + \gamma \max_{a'} Q(s', a') - Q(s, a) \right]
\end{equation*}
% \end{small}
where $r$ is the immediate reward after taking action $a$ in state $s$. This update procedure enables the agent to incrementally learn from every single $(s,a,s',r)$ sample.

Model-based RL algorithms learn the world model, including $ \mathcal{R}(s, a) $ and $ \mathcal{T}(s, a, s') $, and then use planning algorithms to compute the action policy. 
Dyna-Q~\cite{Sutton:1991:DIA:122344.122377} is a model-based RL framework, and includes the two primary components of model-free RL (Q-learning) and probabilistic planning (e.g., value iteration). 
The real-world interaction experience is used for two purposes in Dyna-Q: world model learning, and action policy learning. 
Besides, Dyna-Q is able to generate extra (simulation) experience through interacting with the learned world model, which further speeds up the policy learning process. 
We use declarative action knowledge to prevent the Dyna-Q agent from exploring less-relevant states. 
% Generally, model learning can be realized using supervised machine learning methods, and the learned world models can then be used for computing action policies. 

We use Dyna-Q as a building block in the implementation of GDQ, because it is simple and has been widely studied in the RL literature. 
It should be noted that GDQ practitioners can replace Dyna-Q with other model-based RL methods.

\subsection{Automated Planning}
\label{sec:task}

Automated planning methods aim at computing a sequence of actions toward accomplishing complex tasks~\cite{ghallab2016automated}. 
% Task planners are goal-independent and superior in scaling to large state space by leveraging declarative knowledge. 
One has to declaratively encode action knowledge into an automated planner, including actions' preconditions and effects. 
% The programming languages used for encoding such planners are frequently called \emph{action language}. 
Since the development of STRIPS~\cite{fikes1971strips}, many action languages have been developed for formally representing action knowledge. 
The following shows an example of using STRIPS to formulate action \c{stack}, where preconditions include the robot arm holding object \c{X} and object \c{Y} being clear. 
Executing this action causes the hand to be empty, and object \c{Y} not clear anymore. 
\begin{quote}
\begin{footnotesize}
\begin{verbatim}
operator(stack(X,Y),
         Precond [holding(X),clear(Y)],
         Add [handempty,on(X,Y),clear(X)],
         Delete [holding(X),clear(Y)])
\end{verbatim}
\end{footnotesize}
\end{quote}

There are a number of action languages that can be used for encoding action knowledge, including PDDL that has been widely used in real-world planning systems~\cite{mcdermott1998pddl}.
% PDDL was initially developed for the International Planning Competition (IPC), and has been maintained by the IPC community~\cite{mcdermott1998pddl}. 
Answer set programming (ASP) is a general-purpose knowledge representation and reasoning paradigm, and supports automated planning~\cite{lifschitz2019answer}. 
We use ASP in this research mainly because of its good performance in knowledge-intensive domains~\cite{jiang2019comparison}. 
For instance, in navigation tasks, the robot needs to reason about potentially many rooms and their connections. 
The action knowledge we use is simple and publicly available~\cite{yang2014planning,jiang2019comparison}, so we do not discuss knowledge acquisition in this paper. 

\section{Algorithm}
\label{sec:alg}

In this section, we present Guided Dyna-Q (GDQ), the key contribution of this research. 
GDQ leverages goal-independent action knowledge for sample-efficient policy learning by enabling the interplay between a model-based reinforcement learner and an automated planner. 

We use $ \Pi(\mathcal{S}^A, \mathcal{A}^A, M) $ to represent our automated planner, where $ \mathcal{S}^A $ and $ \mathcal{A}^A $ are the state and action sets respectively. 
% For \textit{Planner}, let $ \Pi(\mathcal{S}^t, \mathcal{A}^t) $ be an ASP program that represented by $ \mathcal{S}^t $ and $ \mathcal{A}^t $ as a set of states and actions respectively.
A task to the automated planner is defined as $M=(s^A_0, s^A_G)$
where $s^A_0, s^A_G \in \mathcal{S}^A $ are the initial state and goal states respectively. 
For simplification, the goal is defined as a single state, though it can be a set of states in practice. 
Given task $M$, an automated planning system
% (e.g.,~\citeauthor{lifschitz2002answer}, \citeyear{lifschitz2002answer}) 
can use $ \Pi(\mathcal{S}^A, \mathcal{A}^A, M) $ to compute a set of plans, $ \mathcal{H} $, where $p \in \mathcal{H}$ is in the form of a sequence of state-action pairs, and each action sequence leads state transitions from the initial state $ s^A_0 $ all the way to the goal state $ s^A_G $. 
\begin{equation*}
    % \label{eq:trajectory}
    p = \left< \left<s^A_0, a^A_0 \right>, \left<s^A_1, a^A_1 \right>, \cdots , \left<s^A_{G}\right>\right>
\end{equation*}

We use an automated planning system~\cite{gebser2011potassco} that supports ``optimisation statements'' for generating a set of the shortest plans (action costs are not considered in this process). 
In order to enable the reinforcement learner to exploit the plans, we introduce a mapping function, $ \mathcal{O} $, that constructs the correspondence between the planner's action space $ \mathcal{A}^A $ and the learner's action space $ \mathcal{A} $, and the correspondence between their state spaces $ \mathcal{S}^A $ and $ \mathcal{S} $. 
\begin{equation*}
    \forall s^A \!\!\in\! \mathcal{S}^A,~ \forall a^A \!\!\in\! \mathcal{A}^A(s^A),~ s \!\in\! \mathcal{O}^{-1}(s^A) \Rightarrow s \!\in\! \mathcal{S},~ a \!\in\! \mathcal{A}(s)
    % \forall s^A \in \mathcal{S}^A,~ a \in \mathcal{A}^A(s^A) \Rightarrow s \in \mathcal{O}^{-1}(s^A),~ s \in \mathcal{S},~ a \in \mathcal{A}(s)
\end{equation*}
where $ \mathcal{A}(s) \subseteq \mathcal{A} $, and $a\in \mathcal{A}(s)$ is applicable in state $ s $.

Next, we describe how the plans generated by the automated planner are used for \emph{optimistic initialization} (Section~\ref{sec:opt}), \emph{policy update} (Section~\ref{sec:update}), and their integration, i.e., the GDQ algorithm.

\begin{algorithm}[t] \footnotesize
\caption{Optimistic Initialization: \textsc{OptInit}}
\label{alg:Algorithm1}
\textbf{Input}: $ \mathcal{S} $, $ \mathcal{A} $, $ \mathcal{S}^A $, $ \mathcal{A}^A $, $ M = (s^A_0, s^A_G) $\\
\textbf{Parameter}: $\gamma$, $\alpha$, $R_{max}$ \\
\textbf{Output}: $ \pi $
\begin{algorithmic}[1] %[1] enables line numbers
    % \STATE // Initialize Q-values and $\mathcal{R}, \mathcal{T}$ of the model.
    \FOR[Initialize Q-values, $\mathcal{R}$, and $\mathcal{T}$]  {$\forall s \in \mathcal{S}, a \in \mathcal{A}(s)$}
    \label{line:init_for_start}
        \STATE $ Q(s, a) \leftarrow 0.0 $; $ \mathcal{R}(s, a) \leftarrow 0.0 $; $ \mathcal{T}(s, a, s_0) \leftarrow 1.0 $ 
    \ENDFOR
    \label{line:init_for_end}
    % \STATE // Request plans to \textit{Planner}.
    
    \STATE $\mathcal{H} \leftarrow \Pi(\mathcal{S}^A, \mathcal{A}^A, M) $
    \COMMENT{Compute a set of optimistic plans}
    \label{line:plans}
    \FOR {$p$ in $\mathcal{H}$}
    \label{line:plan_start}
        \FOR{$\left<s^A, a^A\right>$ in $p$}
            \STATE {$s, a \leftarrow \mathcal{O}(s^A)$}
            \STATE $ \mathcal{R}(s, a) \leftarrow R_{max} $
            \COMMENT{Optimistically update $\mathcal{R}$}
        \ENDFOR
    \ENDFOR
    \label{line:plan_end}
    \STATE $ \pi \leftarrow \text{random policy}$
    \COMMENT {Randomly initialize a policy}
  %  \STATE $ \pi' \leftarrow \pi $
    \WHILE{$\pi \neq \pi'$}
    \label{line:while_start}
        \STATE $ \pi' \leftarrow \pi $
        \FOR{$\forall s \in \mathcal{S} $}
            \STATE $ Q(s,\! \pi(s))\!\! \leftarrow \!\! \mathcal{R}(s,\! \pi(s)) \! + \! \gamma \!\!\! \sum\limits_{s' \in \mathcal{S}} \!\! \mathcal{T}(s,\! \pi(s),\! s')Q(s'\!,\! \pi(s'))$ 
            \STATE $\pi(s) \leftarrow \text{argmax}_a Q(s, a)$
        \ENDFOR
    \ENDWHILE   
    \label{line:while_end}
    \STATE \textbf{return} $\pi$
    \label{line:end}
\end{algorithmic}
\end{algorithm}

\subsection{Optimistic Initialization}
\label{sec:opt}

The plans computed by the automated planner are referred to as \emph{optimistic} plans, because real-world domain uncertainty is frequently overlooked in building the planners. 
For instance, a robot taking the action of ``\emph{navigate to room R}'' sometimes does not result in the robot being in \emph{room R} due to the possibility of obstacles blocking the way. 
The goal of \emph{optimistic initialization} (\textsc{OptInit}) is to use the plans computed by the automated planner to initialize Q-values, and prevent the agent from exploring less-relevant states. 

Algorithm~\ref{alg:Algorithm1} presents our optimistic initialization process.
The input includes the state and action spaces of both the reinforcement learner and the automated planner. 
$M$ is the provided task.
The output is an initial policy $\pi$, which is generated by the agent interacting with the world.

Lines~\ref{line:init_for_start}-\ref{line:init_for_end} are used for initializing the Q-values, as well as the transition and reward functions. 
% \textit{Learner} initialize the Q-value, the transition function $ \mathcal{T} $, and the reward function $ \mathcal{R} $ with unit values, respectively.
The transition function is initialized in a way that all state-action pairs deterministically lead to the initial state. 
This setting is necessary because all plans computed by the automated planner start from the initial state. 
% $ Q_0(\cdot, \cdot) = 0.0$, , and $ \mathcal{R}(\cdot, \cdot) = 0.0 $ in this paper.
Given task $ M=(s^A_0, s^A_G) $, our automated planner computes a set of optimistic plans in Line~\ref{line:plans}. 
The two for-loops in Lines~\ref{line:plan_start}-\ref{line:plan_end} assign the highest reward, $R_{max}$, to the reward of all state-action pairs that appear in the plans from our automated planner. 
This is similar to how R-MAX realizes the trade-off between exploration and exploitation~\cite{brafman2002r}. 
% Then updates the reward function only of state-action sets which are included in the trajectories as transitions using $ R_{max} $.
Lines~\ref{line:while_start}-\ref{line:while_end} are used for computing an action policy using policy iteration.
Finally, the computed policy is returned in Line~\ref{line:end}.

It should be noted that the agent has not started interacting with the environment while running \textsc{OptInit} (Algorithm~\ref{alg:Algorithm1}). 
This initialization process enables the agent to prioritize states that are more relevant to the current task when exploring its working environment, which enables the agent to accomplish tasks more efficiently.  

\begin{algorithm}[t]
\small
\caption{Policy Update: \textsc{PolicyUp}
% \commentsa{None of the three algorithms are reflecting any exploration at all. Line 2 should be better expressed}
}
\label{alg:Algorithm2}
\textbf{Input}: $ \mathcal{S} $, $ \mathcal{A} $, $ \mathcal{S}^A $, $ \mathcal{A}^A $, $ M = (s^A_0, s^A_G) $, $\pi$, $C$, $R_{sum}$\\
\textbf{Parameter}: $\gamma$, $\alpha$, $R_{max}$, $ m $, $N$\\
\textbf{Output}: $\mathcal{R}, \mathcal{T}, \pi$
\begin{algorithmic}[1] %[1] enables line numbers
    \STATE Collect the current world state $s$ from the world
    \label{line:state}
    \STATE $a \leftarrow $ $\pi(s)$, with $\epsilon$ exploration rate
    % \COMMENT{Select an action using policy $\pi$}
    \STATE Collect resulting state $s'$ and reward $r$ after taking $a$
    \STATE Update the Q-value using real interaction experience\\
    $Q(s, a) \leftarrow Q(s, a) + \alpha[r + \gamma \max_{a'} Q(s', a')-Q(s, a)]$ \label{line:update}
    \STATE $ C(s, a, s') \leftarrow C(s, a, s') + 1 $
    \STATE $ R_{sum}(s, a) \leftarrow R_{sum}(s, a) + r $ 
    \IF{$\sum_{s'}{C(s, a, s')} > m$}
    \label{line:if}
        % \STATE $ \mathcal{T}(s, a, s') \leftarrow \frac{C(s, a, s')}{\sum_{s'}{C(s, a, s')}} $
        \STATE $ \mathcal{T}(s, a, s') \leftarrow C(s, a, s') / \sum_{s''}{C(s, a, s'')} $
        % \STATE $ \mathcal{R}(s, a) \leftarrow \frac{\mathcal{R}(s, a) + r}{\sum_{s'}{C(s, a, s')}} $
        \STATE $ \mathcal{R}(s, a) \leftarrow R_{sum}(s, a) / \sum_{s''}{C(s, a, s'')} $
    \ENDIF
    \STATE $ M \leftarrow (\mathcal{O}(s), s^A_G) $
    \STATE $ \mathcal{H} \leftarrow \Pi(\mathcal{S}^A, \mathcal{A}^A, M) $ 
    \COMMENT{Compute a set of plans}
    \label{line:plan2}
    \FOR{$n$ in $\{1 \cdots N\}$}
    \label{line:for_start2}
        \STATE $ p \leftarrow \text{randomly selected plan in } \mathcal{H} $ 
        \STATE$ \left<s^A, a^A\right> \leftarrow \text{randomly selected transition in } p $
        \STATE {$s, a \leftarrow \mathcal{O}^{-1}(s^A)$}
        % \comments{Same comment as in Alg 1. Not sure how ``applying'' it is changing anything in the algorithm. }
        % \STATE $ Q(s, a) \leftarrow (1-\alpha) Q(s,a) + \alpha[\mathcal{R}(s, a) + \gamma \max_{a'} Q(s', a')] $
        \STATE {Update $Q(s,a)$ using Bellman equation. }
    \ENDFOR
    \label{line:for_end2}
    \STATE $\forall s\in \mathcal{S}, \forall a\in \mathcal{A}, \pi(s) \leftarrow \text{argmax}_a Q(s, a)$
    \STATE \textbf{return} $\mathcal{R}, \mathcal{T}, \pi$
    \label{line:return}
\end{algorithmic}
\end{algorithm}

\subsection{Policy Update and GDQ}
\label{sec:update}

The previous subsection (Section~\ref{sec:opt}) presents the process of initializing the Q-function using the optimistic plans generated by our automated planner. 
Given the initialized $Q$-value function, the agent is able to compute an initial policy, and use this policy to interact with the real world. 
This subsection describes how the interaction experience, along with the automated planner, is used to update the $Q$-value function at runtime. 
Intuitively, the automated planner serves as an optimistic simulator to enable the reinforcement learner to learn from interaction experience in simulation. 

% Second, \textit{Learner} learns the model of the domain while updating Q-values through experiences interacting both with the real environment and the learned model. 
% When \textit{Learner} planning using the world model, \textit{Planner} refines the plans of simulated experience by exploiting action knowledge.

Algorithm~\ref{alg:Algorithm2} presents the 
% runtime 
policy update process. 
% Its input is the same as Algorithm~\ref{alg:Algorithm1}, except that it also includes an action policy and a counter function $C$. 
Its input includes an action policy, a counter function $C$, and a reward counter function $R_{sum}$, in addition to the input of Algorithm~\ref{alg:Algorithm1}.
This policy is provided by Algorithm~\ref{alg:Algorithm1}. 
Parameter $m$ is a threshold, representing how many times a state-action pair has been selected. 
Parameter $N$ is used for determining how many state-action pairs are simulated using the automated planner. 
The output includes not only a policy, but also the reward and transition functions, because our reinforcement learner is model-based.

% To learn a policy, \textit{Learner} updates the Q-values while \textit{Learner} learns the model.
% The algorithm to learn a world model and to update policy is shown in Algorithm \ref{alg:Algorithm2}.
Lines~\ref{line:state}-\ref{line:update} are used for interacting with the real world using the current action policy, $\pi$. 
% \textit{Learner} updates the Q-value by using direct RL through a real experience after the agent takes action $ a $ following the current policy $ \pi $ in the state $ s $.
Then, $ C(s, a, s') $ is increased by one for updating the number of state-action-state triples. 
If the agent has visited a state-action pair for more than $m$ times (Line~\ref{line:if}), the transition and reward functions are updated. 
% a counter of a transition $ (s, a, s') $, 
Intuitively, $ m $ is a parameter that indicates a state-action pair being \emph{known} or \emph{unknown}. 
% Observing a next state $ s' $ and a immediate reward $ r $, \textit{Learner} updates the model using $ C(s, a, s') $ if the total number of times a state-action set is visited is over $ m $. 
% \textit{Learner} further updates the Q-value by using the model updated in the previous line.
In Line~\ref{line:plan2}, the automated planner generates a set of plans. 
% and the plans are updated model and plans generated by \textit{Planner}.
% \textit{Planner} generates all of the plans that the paths start from the current state $ s $ to the target state $ s_G $. 
Using the generated plans, we randomly select one transition from a randomly-selected plan $ p \in \mathcal{H} $, and update the Q-value accordingly.
% {\color{red}We need to add why we do not update Q-values for all sets of states and actions.}
This Q-value update process is repeated for $N$ times in Lines~\ref{line:for_start2}-\ref{line:for_end2}. 
Finally, the reward function, transition function, and updated policy are returned. 

Algorithm~\ref{alg:gdq} is simply an integration of the two sub-procedures for optimistic initialization (Algorithm~\ref{alg:Algorithm1}) and repeatedly conducted runtime policy update (Algorithm~\ref{alg:Algorithm2}), which identifies the main contribution of this research. 
Informally, Algorithm~\ref{alg:Algorithm1} helps the agent avoid the near-random exploration behaviors through a ``warm start'' enabled by our automated planner, and Algorithm~\ref{alg:Algorithm2} guides the agent to only try the actions (in the real world) that can potentially lead to the ultimate goal. 
Next, we present an instantiation of GDQ followed by our experiment setup, and experimental results from comparisons between GDQ and a number of baseline methods selected from the literature.

% \begin{figure}[t]
%     \centering
%     \includegraphics[width=0.4\textwidth]{images/REWARD_env1S0G17E500I25}\\ 
%     \includegraphics[width=0.4\textwidth]{images/REWARD_env1S14G17E500I25}
%     \caption{Average cumulative reward over 10 runs (each run includes a row of 500 episodes), while the robot working on Task $A$ (Top) and Task $B$  (Bottom). }
%     \label{fig:exp12}
% \end{figure}

% \begin{figure}[t]
%     \centering
%     \includegraphics[width=0.4\textwidth]{images/REWARD_env1S0G18E500I25} \\
%     \includegraphics[width=0.4\textwidth]{images/REWARD_env1S14G18E500I25}
%     \caption{Average cumulative reward over 10 runs (each run includes a row of 500 episodes), while the robot working on Task $C$ (Top) and Task $D$  (Bottom). }
%     \label{fig:exp34}
% \end{figure}

\begin{algorithm}[t]
\small
\caption{Guided Dyna-Q (GDQ)} 

\label{alg:gdq}
\textbf{Input}: $ \mathcal{S} $, $ \mathcal{A} $, $ \mathcal{S}^A $, $ \mathcal{A}^A $, $ M = (s^A_0, s^A_G) $, $\pi$, $C$\\
% \textbf{Parameter}: $\gamma$, $\alpha$, $R_{max}$, $ m $, $N$\\
\textbf{Output}: $\pi$
\begin{algorithmic}[1] %[1] enables line numbers
    % \WHILE{A task has been received}
        \STATE Call Algorithm-1: 
        $\pi \leftarrow \textsc{OptInit}(\mathcal{S}, \mathcal{A}, \mathcal{S}^A, \mathcal{A}^A, M)$
        % \STATE $\pi \leftarrow \textsc{OptInit}(\mathcal{S}, \mathcal{A}, \mathcal{S}^A, \mathcal{A}^A,  M)$
        \STATE $\forall s\in \mathcal{S}, \forall a\in \mathcal{A}, \forall s'\in \mathcal{S}, C(s,a,s') \leftarrow 0, R_{sum}(s, a) \leftarrow 0$ 
        \WHILE{Current state $s$ is \textbf{not} terminal}
            \STATE Call Algorithm-2: 
            $$\mathcal{R}, \mathcal{T}, \pi \leftarrow \textsc{PolicyUp}(\mathcal{S}, \mathcal{A}, \mathcal{S}^A, \mathcal{A}^A,  M, \pi, C, R_{sum})$$ 
        \ENDWHILE
        \STATE \textbf{return} $\pi$
    % \ENDWHILE
\end{algorithmic}
\end{algorithm}

\subsection{GDQ Instantiation}

While GDQ is a general-purpose algorithm for knowledge-based RL, its implementation requires a task planner that is domain-dependent (still task-independent). 
We consider a mobile robot navigation domain, where the robot needs to navigate in an indoor office environment. 
The rooms, including hallways, are connected through doors that are of different sizes. 
Big doors are more friendly to the robot, though the robot needs to learn the ``success rate'' of navigating through doors from trial and error. 
Our robot does not have an arm and hence needs human help to open doors. 
Some doors are located in areas where human help is better available, while in some areas the robot might have to wait a long time until people show up to help. 

In each trial (episode), the robot is tasked with navigating from its initial position to a goal position. 
There are doors connecting rooms and corridors, and there are different costs and success rates in navigation and door opening actions. 
The robot has four types of actions of $\texttt{goto(P,I)}$, $\texttt{approach(D,I)}$, $\texttt{opendoor(D,I)}$, and $\texttt{gothrough(D,I)}$ for navigational purposes, where $\texttt{P}$ is one of the 19 positions, $\texttt{D}$ is one of the 6 doors, and $\texttt{I}$ is the step number. 
% Each type of action has a number of instances, e.g., \textit{gothrough} can be realized using any of the six doors. 
The actions and parameters together form a large action space for our RL agent. 
As an example, the following rule defines the effect of action $\texttt{\small gothrough(D,I)}$,  
\begin{quote}
\begin{footnotesize}
\begin{verbatim}
at(R2,I+1):- gothrough(D,I), 
            at(R1,I), acc(R1,D,R2), I<n.
\end{verbatim}
\end{footnotesize}
\end{quote}
where $\texttt{\small acc(R1,D,R2)}$ indicates that rooms $\texttt{\small R1}$ and $\texttt{\small R2}$ are connected through door $\texttt{\small D}$. 
The rule states that, if the robot is at room $\texttt{\small R1}$ at timestamp $\texttt{\small I}$, and $\texttt{\small acc(R1,D,R2)}$ is true, then going through door $\texttt{\small D}$ causes the robot to be in room $\texttt{\small R2}$ in the next step. 
$\texttt{\small n}$ is the maximum steps allowed in the navigation task.\footnote{More information on ASP-based planning systems is available online: ``https://github.com/potassco/guide''.
The code of GDQ is available at ``https://github.com/YoheiHayamizu/gdq''}
% We implement GDQ algorithm~\footnote{We will open-source after acceptance} on an existing codebase~\footnote{We will cite it after paper acceptance} for our robot navigation and task completion. 

We used the BWI codebase~\cite{khandelwal2017bwibots}, including their code for automated planning, in our robot navigation experiments. 
GDQ is a general framework that can be realized using different building blocks for learning and planning. 
For instance, the Dyna-Q component for model-based RL can be replaced by MBMF~\cite{bansal2017mbmf} or I2A~\cite{racaniere2017imagination}; the ASP-based automated planner can be constructed using STRIPS~\cite{fikes1971strips} or PDDL~\cite{mcdermott1998pddl}.

\begin{figure*}[ht]
  \centering
        \includegraphics[height=3.7cm]{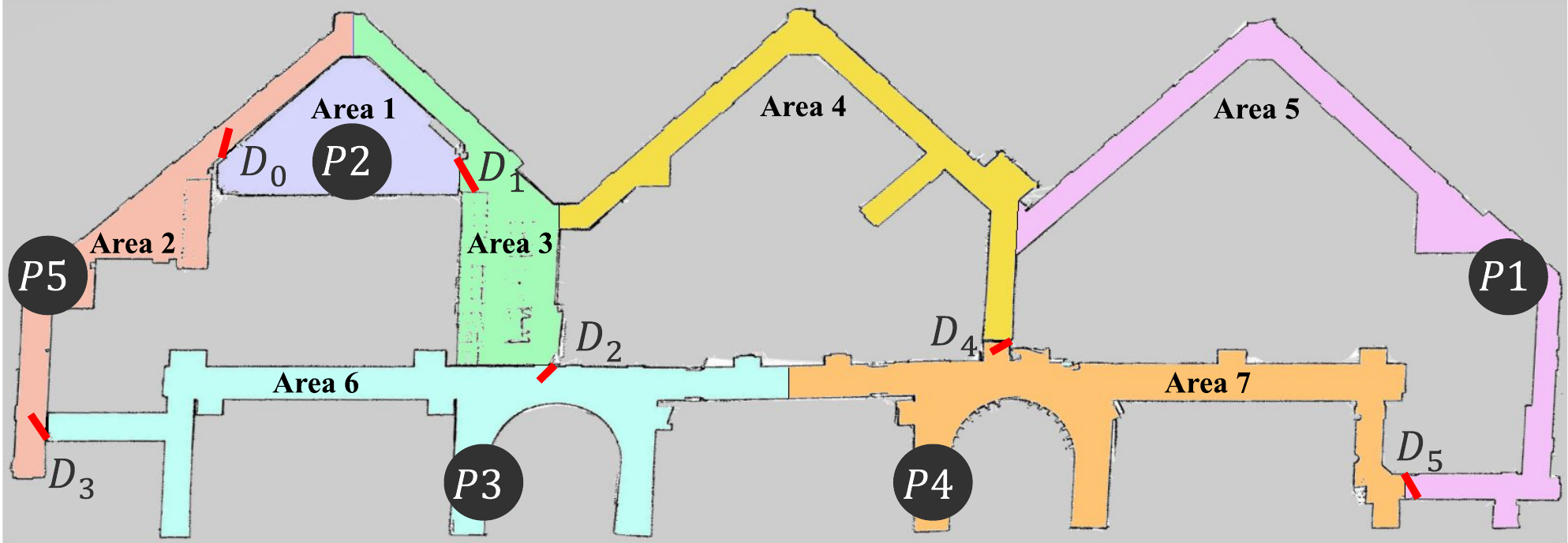}
        \includegraphics[height=3.7cm]{images/robot_front.pdf}
        \includegraphics[height=3.7cm]{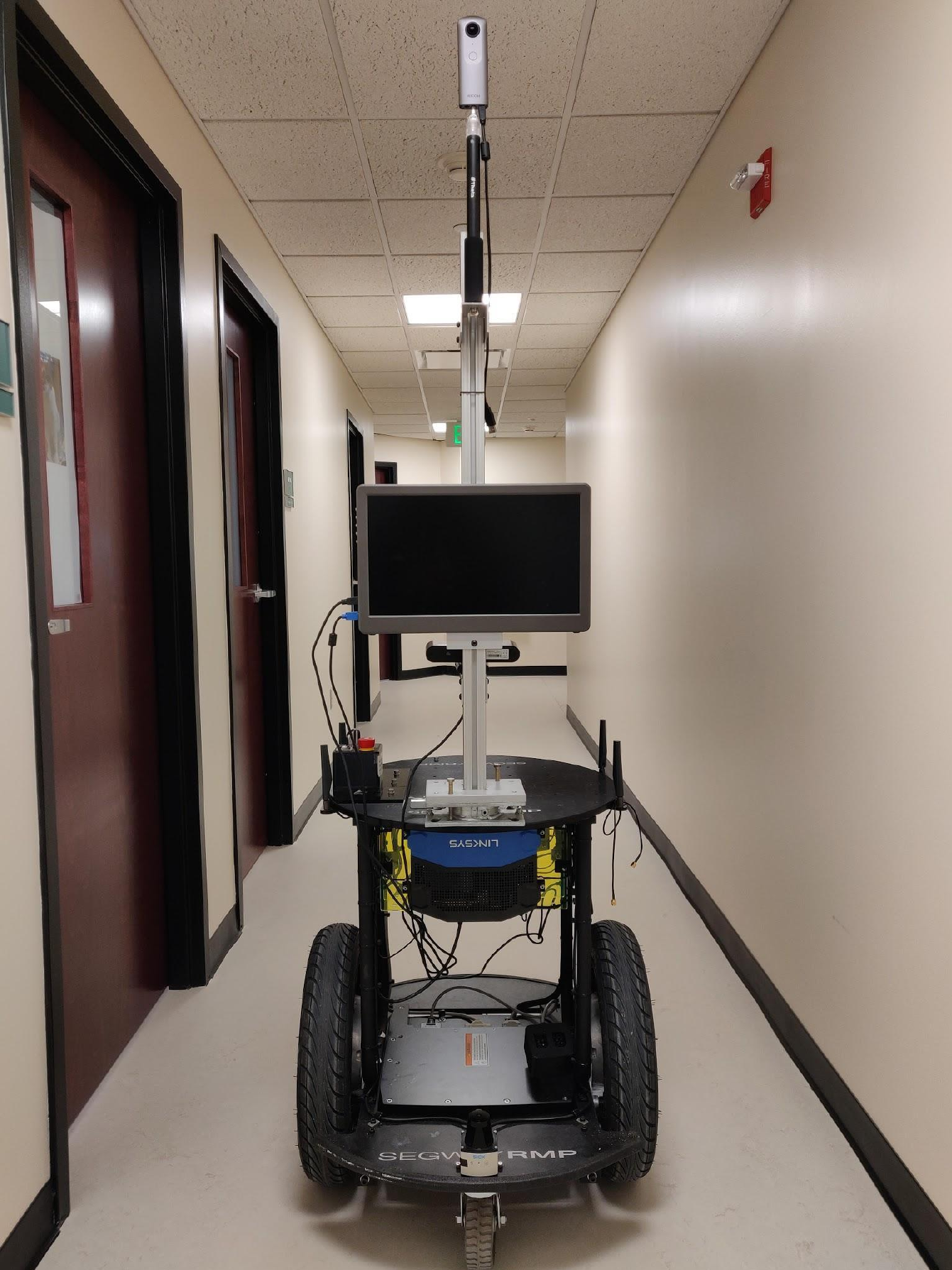}
        \label{fig:segbot}
    \caption{An occupancy-grid map (\textbf{Left}) of an indoor office environment with more than 20 rooms, where the map was built using a mobile robot running simultaneous localization and mapping (SLAM) algorithms~\cite{montemerlo2002fastslam,thrun2005probabilistic} and each pixel is labeled in color with its semantic meaning (area number). 
    The initial and goal positions used in the experiments are named as $P1...P5$. 
    Red lines refer to the room doors; and the front (\textbf{Middle}) and back (\textbf{Right}) of our Segway-based mobile robot platform that was used for building the map and evaluations of the GDQ algorithm. 
    }
    % https://docs.google.com/drawings/d/1GLZYtgO5703p0-B7lbbIrbi9aanE0XW2-1q05On9GxQ/edit?usp=sharing
    \label{fig:map}
\end{figure*}

\section{Experiment}
\label{sec:exp}

In this section, we focus on experimentally evaluating the following three hypotheses that GDQ: 
\begin{enumerate}
    \item Performs better than existing RL methods from the literature in cumulative reward (\textbf{Hypothesis-I}); 
    \item Enables the robot to reduce the number of visits to ``irrelevant'' areas (\textbf{Hypothesis-II}), where an area is deemed \textbf{relevant} to a navigation task, if there exists one optimal plan that requires the robot navigating that area; and 
    \item Is more robust to goal changes (\textbf{Hypothesis-III}).
\end{enumerate}

GDQ has been compared with a model-free RL baseline (Q-Learning), a model-based RL baseline (Dyna-Q), and a knowledge-based RL approach called DARLING~\cite{DBLP:journals/ai/LeonettiIS16} that reasons with action knowledge to avoid ``unreasonable'' exploratory behaviors.
% in the real world

%In order to evaluate Hypothesis-II, 
We define seven areas in the map as shown in Figure~\ref{fig:map}, and each area was manually separated into four subareas (each corresponds to a state). 
Some of the areas are directly accessible to each other (e.g., Areas 6 and 7), whereas the others are connected through doors (e.g., Areas 1 and 2). 
We have labeled six doors in the map that our robot can use to enter rooms. 
All doors are automatic, meaning that, to go through a door, the robot must get close to it, and open it before taking the \emph{gothrough} action.
The real robot needs help from people for door opening actions (printing on its screen ``\emph{Please help me open the door}''), which requires different time durations depending on people's availability. 
In simulation, each door is associated with a success-rate distribution, and another distribution over action costs. 
We tried to give realistic distributions to match the real door's physical properties (width, location, weight, etc). 
$D0$, $D2$, and $D5$ are difficult doors, where $D2$ is the most difficult to be opened. 
$D1$, $D3$, and $D4$ are easy, where $D3$ is the easiest. 
The simulation environment used in experiments has been created as an extension of OpenAI Gym, a standard platform for RL research~\cite{brockman2016openai}.

\begin{figure*}[!ht]
     \centering
     \begin{subfigure}[b]{0.37\textwidth}
         \centering
         \includegraphics[width=\textwidth]{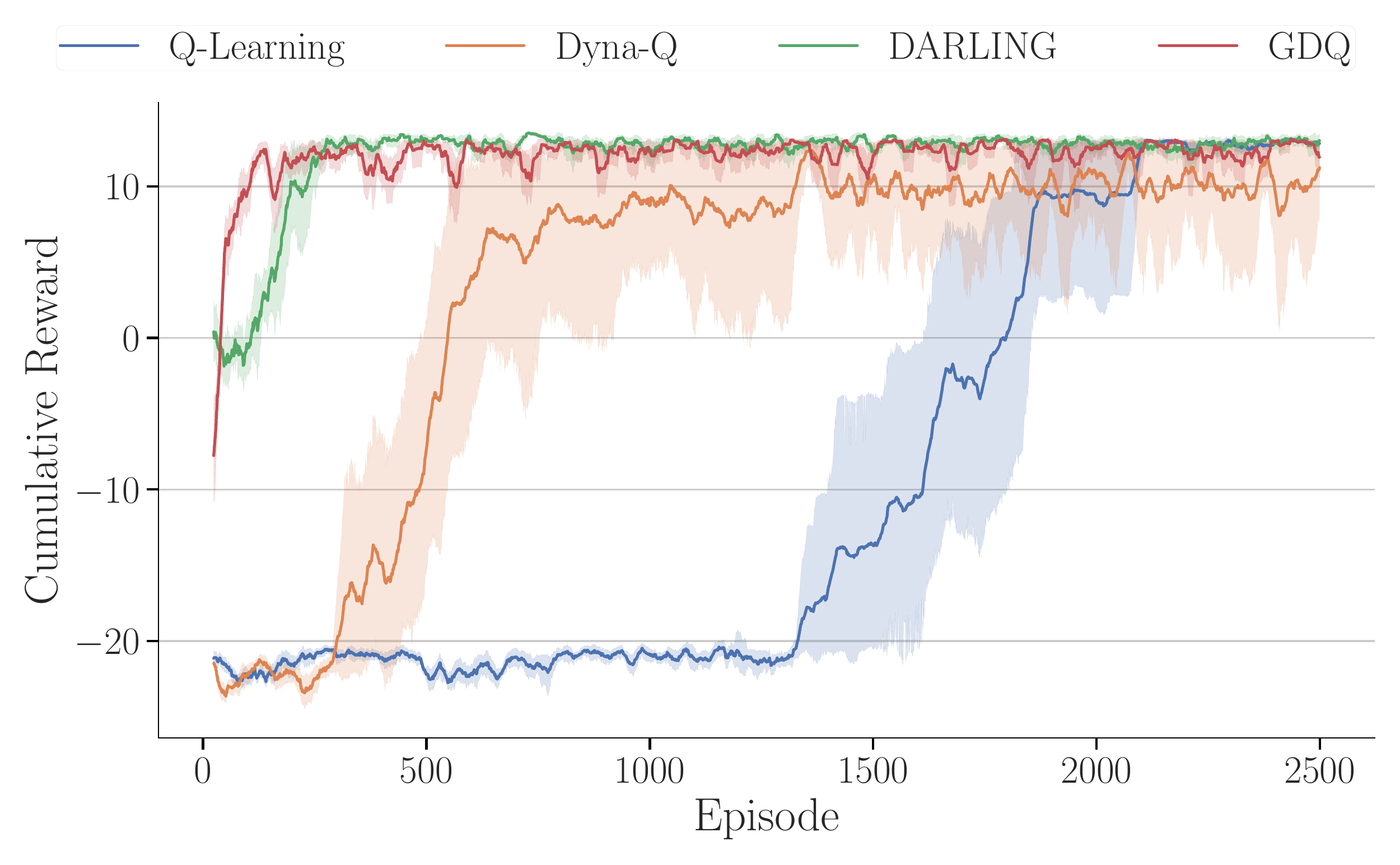}
         \caption{Task $A$: $ M_A = (P1, P3)$}
         \label{fig:exp1}
     \end{subfigure}
     \begin{subfigure}[b]{0.37\textwidth}
         \centering
         \includegraphics[width=\textwidth]{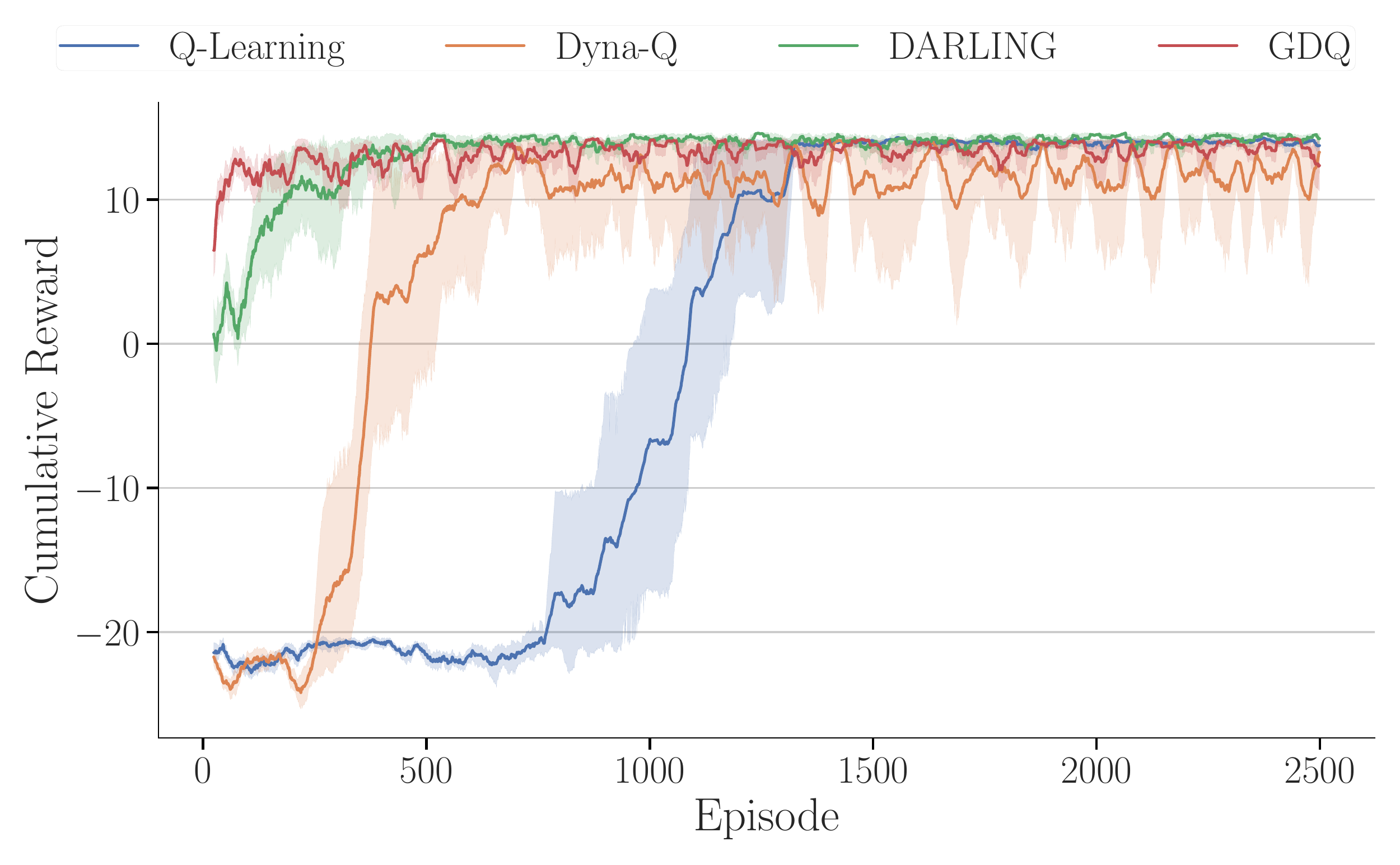}
         \caption{Task $B$: $M_B = (P1, P4) $}
         \label{fig:exp2}
     \end{subfigure}
     \\
     \begin{subfigure}[b]{0.37\textwidth}
         \centering
         \includegraphics[width=\textwidth]{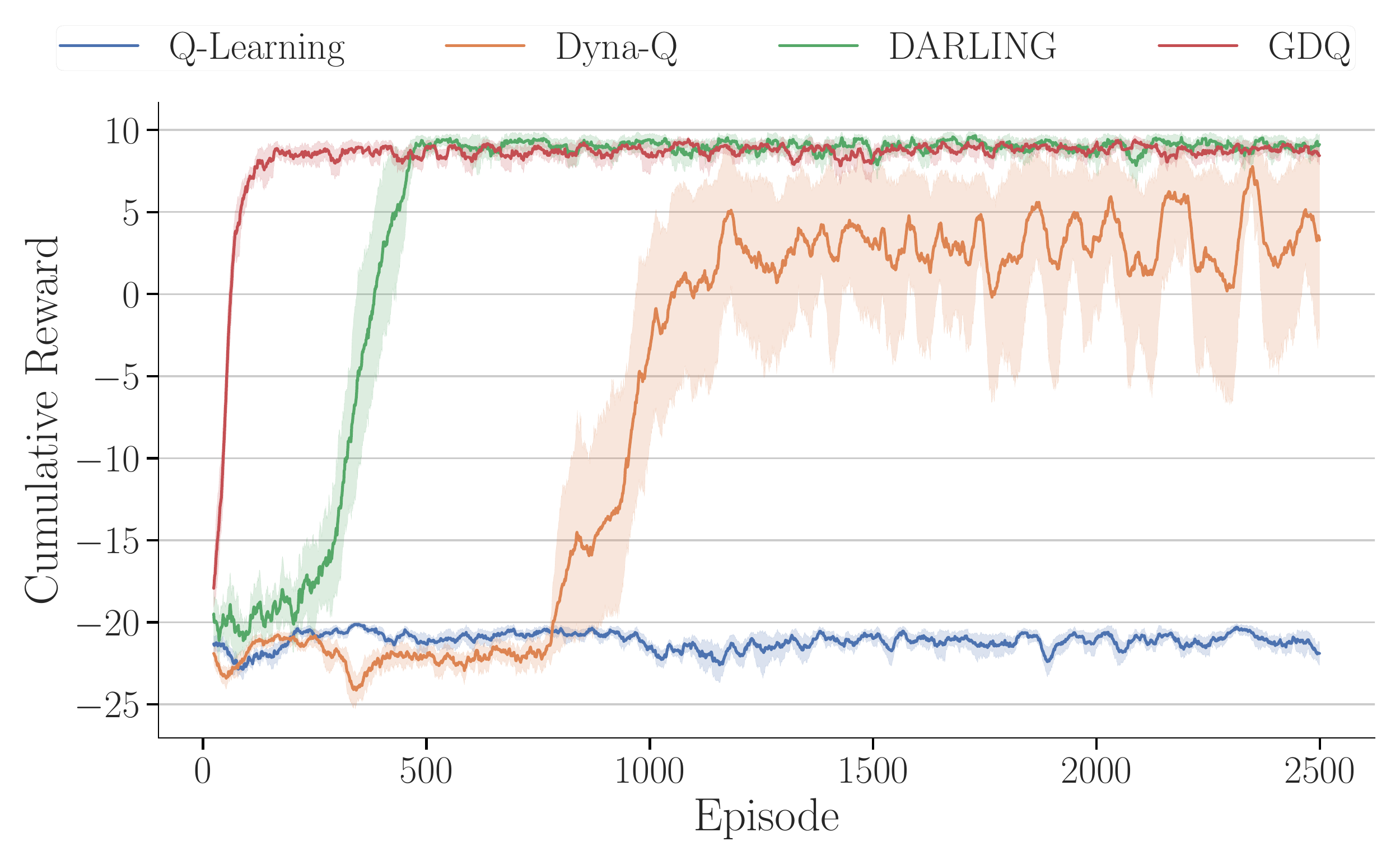}
         \caption{Task $C$: $ M_C = (P2, P3)$}
         \label{fig:exp3}
     \end{subfigure}
     \begin{subfigure}[b]{0.37\textwidth}
         \centering
         \includegraphics[width=\textwidth]{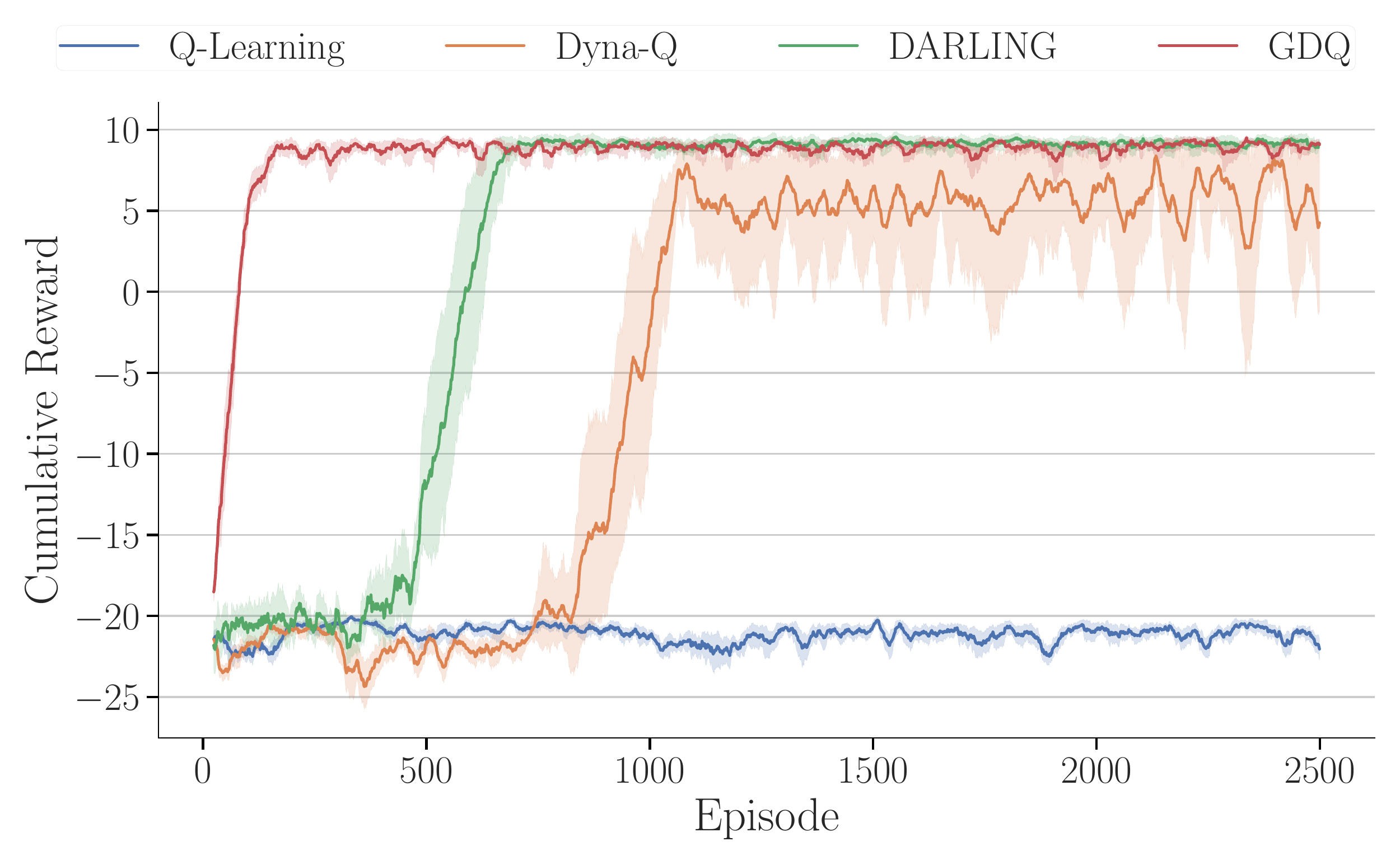}
         \caption{Task $D$: $ M_D = (P2, P4)$}
         \label{fig:exp4}
     \end{subfigure}
     \caption{Average cumulative reward over ten runs (each run includes 2500 episodes), while the robot working on different indoor navigation tasks in simulation. 
     GDQ produced the best performance in all four tasks. 
     }
     \label{fig:exp1234}
\end{figure*}

\subsection{Simulation Experiment}
\label{sec:sim}

The agent receives a big reward, $R_{max}$, in successful trials; receives a big penalty, $-R_{max}$, in failure trials; and receives a small cost, $c$, at all other times. 
In this experiment, $ R_{max} = 20$, and $ c = 1$. 
Our agent tries a random action in probability $\epsilon=0.1$.
The learning rate is $\alpha=0.1$, and the discount factor is $\gamma = 0.95$. 
We set a threshold as the maximum number of actions allowed in each episode: not being able to complete a task within 20 actions makes a trial unsuccessful. 
Each ``run'' includes $2500$ episodes in a row, and each data point of our figures is an average of over $10$ runs. 
We have conducted four independent experiments in simulation.

\paragraph{Cumulative Reward}
Figures~\ref{fig:exp1234} presents the cumulative rewards collected from the robot conducting Tasks $A$, $B$, $C$, and $D$, as well as the tasks' initial and goal positions. 
We observe that GDQ performed the best in learning rate in comparison to the four baselines, which supports Hypothesis-I.  

Looking into Task-$C$ (bottom-left subfigure), there are the following valid routes that can lead to the goal position while producing different costs and success-rates: 
$ [1 \rightarrow 2 \rightarrow 6] $, $ [1 \rightarrow 3 \rightarrow 6] $, $ [1 \rightarrow 3 \rightarrow 4 \rightarrow 7 \rightarrow 6] $, $ [1 \rightarrow 3 \rightarrow 2 \rightarrow 6] $, $ [1 \rightarrow 3 \rightarrow 4 \rightarrow 5 \rightarrow 7 \rightarrow 6] $, where each number corresponds to the index of an area. 
The shortest routes are $ [1 \rightarrow 2 \rightarrow 6] $, and $ [1 \rightarrow 3 \rightarrow 6] $. 
However, the two routes have doors of $ D0 $ and $ D2 $, which are both difficult. 
In comparison, $ [1 \rightarrow 3 \rightarrow 2 \rightarrow 6] $ provides the best trade-off between traveling distance and door difficulty, and is the best solution.
GDQ enabled the robot to converge to this solution earlier than all other baseline methods.

\paragraph{Exploration}
Aiming to evaluate Hypothesis-II on exploration, we manually provided the ground truth relevance information, where we introduce function $IRR$ that maps a task to a set of irrelevant areas 
$$ \textit{irrelevant areas} \leftarrow IRR(\textit{task})$$

% Back to our testing domain, the irrelevant areas to each task are: $\{4,5,7\} \leftarrow IRR(A)$, $\{2,5\} \leftarrow IRR(B)$, $\{1,2,3\} \leftarrow IRR(C)$, and $\{1,2,3,6\} \leftarrow IRR(D)$
Back to our testing domain, the irrelevant areas to each task are: $\{1,2,3\} \leftarrow IRR(A)$, $\{1,2,3,6\} \leftarrow IRR(B)$, $\{4,5,7\} \leftarrow IRR(C)$, and $\{2,5\} \leftarrow IRR(D)$
% \begin{itemize}
%     \item $Task\,A$: $ Area\,4 $, $ Area\,5 $, $ Area\,7 $
%     \item $Task\,B$: $ Area\,5 $
%     \item $Task\,C$: $ Area\,1 $, $ Area\,2 $, $ Area\,3 $, $ Area\,4 $
%     \item $Task\,D$: $ Area\,1 $, $ Area\,2 $, $ Area\,3 $, $ Area\,4 $, $ Area\,6 $
% \end{itemize}

Table~\ref{tab:explore} shows the results in evaluating the performances in exploration. 
The bold text indicates the method that produced the least visits, and we say \emph{the robot successfully avoids the area using this method. }
Consider the last four rows that correspond to Task-$D$. 
We see that GDQ enabled the robot to visit Area-2 for as few as only $37.2$ times, which is much lower than the number of visits required by the other methods (say Q-Learning requires $479.8$ visits), while still produced the best performance in policy quality. 
This observation is consistent with our prior knowledge that Areas 2 and 5 are less-relevant to Task-$D$. 
In all four tasks, the robot successfully avoided the irrelevant areas (see the highlighted areas with bold and the listed irrelevant areas in column \emph{Task}), supporting Hypothesis-II.

\begin{figure*}[!ht]
     \centering
     \begin{subfigure}[b]{0.37\textwidth}
         \centering
         \includegraphics[width=\textwidth]{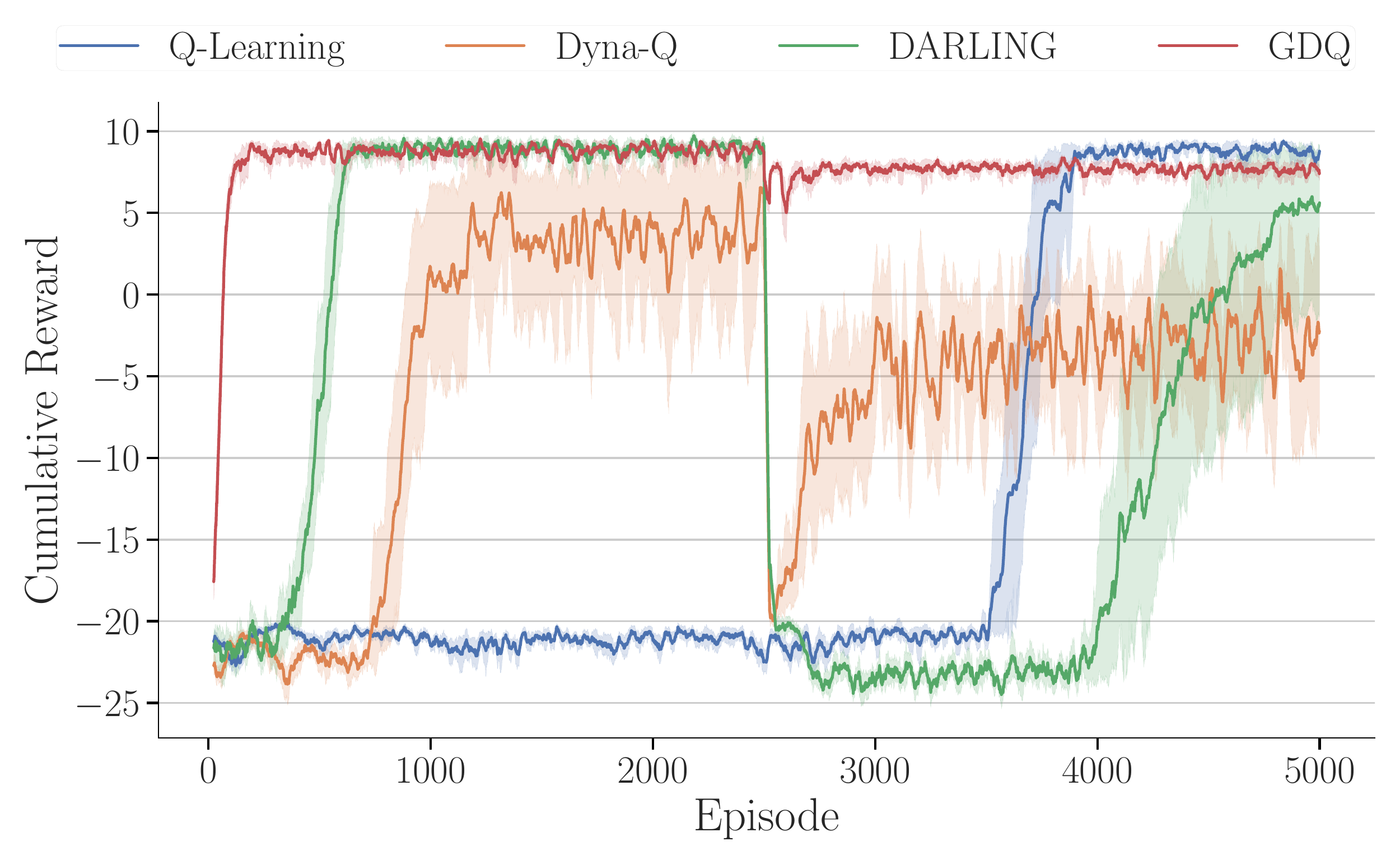}
         \caption{Task $C$ to Task $D$}
         \label{fig:switch1}
     \end{subfigure}
     \begin{subfigure}[b]{0.37\textwidth}
         \centering
         \includegraphics[width=\textwidth]{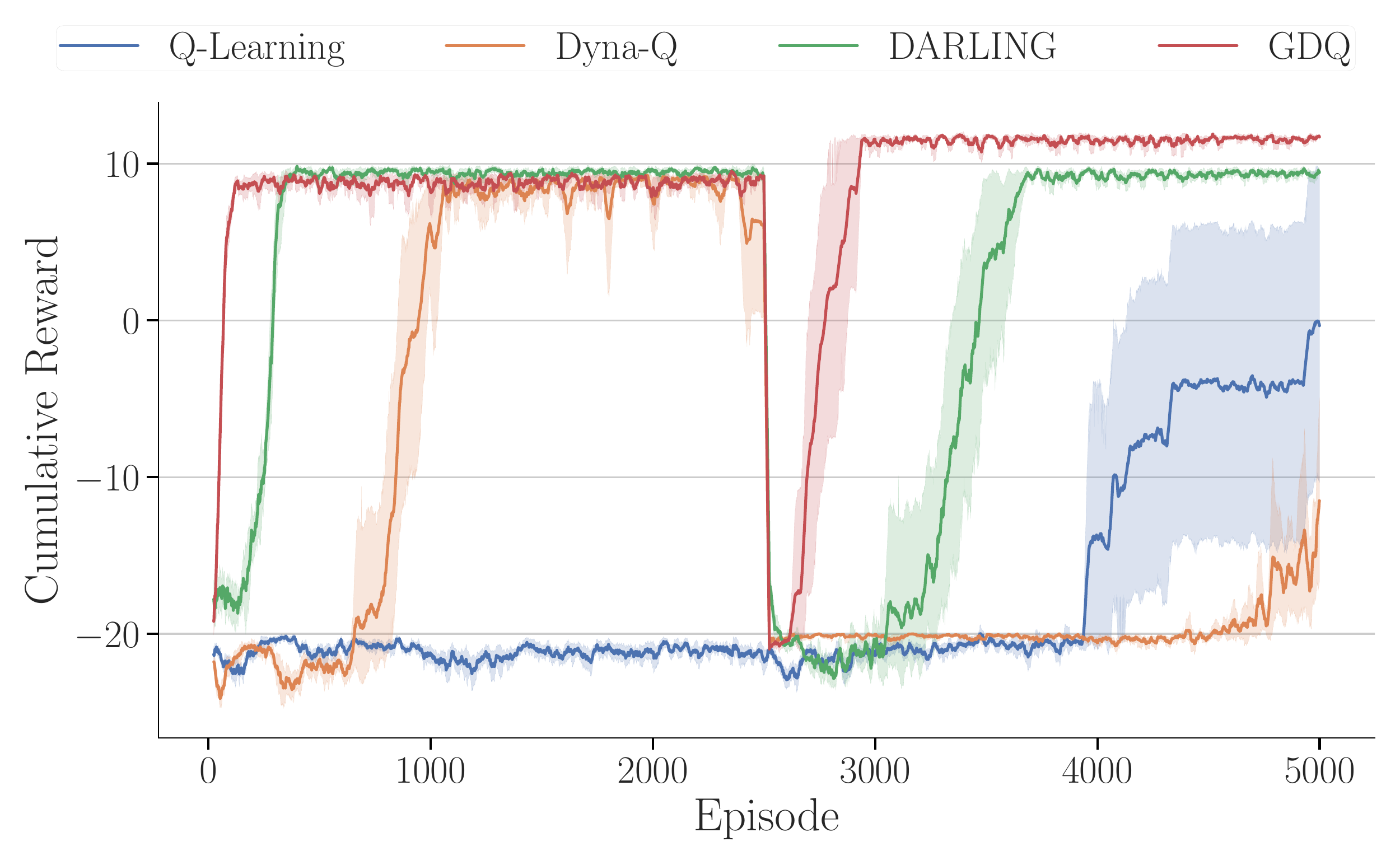}
         \caption{Task $C$ to Task $E$: $ M_E = (P3, P1)$}
         \label{fig:switch4}
     \end{subfigure}  
     \caption{The robot conducted two tasks in each experiment to evaluate Hypothesis-III, where the task was changed at the 2500$th$ episode. 
     Average cumulative reward over ten runs (each run includes 5000 episodes), while the robot working on different indoor navigation tasks in simulation. 
     GDQ produced the best performance in the two experiments.}
     \label{fig:exp_switch}
\end{figure*}

\begin{table*}[ht]
  \centering

  \caption{This table shows how many times the robot visited each area using four different methods (including GDQ) in conducting the four different tasks. The goal is to show GDQ helps the robot avoid visiting areas that are less-relevant to the given task. Bold text indicates the fewest visits to the less-relevant areas (listed in the ``Task'' column) among the four methods. 
  }
  \label{tab:explore}
  \begin{small} 
  \begin{tabular}{c||c||r|r|r|r|r|r|r}
    \toprule
    Task & Method & $Area\,1$ & $Area\,2$ & $Area\,3$ & $Area\,4$ & $ Area\,5 $ & $Area\,6$ & $Area\,7$ \\
    \midrule
    
     & \textit{GDQ} & $ \mathbf{ 0.0~\!(0.0) }$ & $ 14.3~\!(0.6) $ & $ \mathbf{ 23.8~\!(1.1) }$ & $ 944.0~\!(42.5) $ &  $ 351.9~\!(15.9) $ & $ 272.1~\!(12.3) $  & $ 613.2~\!(26.4) $ \\
    Task A & \!\textit{DARLING}\! & $ 1.0~\!(0.1) $ & $ \mathbf{ 10.0~\!(0.5) }$  &  $ 27.9~\!(1.3) $ & $ 853.9~\!(40.8) $ & $ 388.3~\!(18.6) $ & $ 260.1~\!(12.4) $ & $ 551.6~\!(26.4) $ \\
    1, 2, 3 & \textit{Dyna-Q} & $ 12.5~\!(0.4) $ & $ 169.6~\!(5.5) $ & $ 312.3~\!(10.2) $ & $ 1146.2~\!(37.4) $ & $ 668.6~\!(21.8) $ & $ 200.2~\!(6.5) $ & $ 553.1~\!(18.6) $ \\
    & \!\!\textit{Q-Learning}\!\! & $ 16.6~\!(0.4) $ & $ 394.4~\!(9.3) $ & $ 833.3~\!(19.7) $ & $ 1435.7~\!(34.0) $ & $ 1126.3~\!(26.7) $ & $ 93.4~\!(2.2) $ & $ 322.2~\!(7.6) $ \\
    \midrule
  
     & \textit{GDQ} & $ \mathbf{ 0.0~\!(0.0) }$  & $ \mathbf{ 0.5~\!(0.0) }$ & $ \mathbf{ 3.2~\!(0.2) }$ & $ 935.4~\!(50.0) $  & $ 352.3~\!(18.8) $ & $ \mathbf{ 0.0~\!(0.0) }$ & $ 581.4~\!(31.0) $ \\
    Task B & \!\textit{DARLING}\! & $ 0.6~\!(0.0) $ & $ 4.7~\!(0.3) $ & $ 16.4~\!(0.9) $ & $ 839.9~\!(47.7) $ & $ 353.3~\!(20.1) $ & $ 1.2~\!(0.0) $ & $ 545.1~\!(31.0) $ \\
    1,2,3,6 & \textit{Dyna-Q} & $ 0.7~\!(0.0) $ & $ 50.3~\!(1.6) $ & $ 177.0~\!(5.6) $ & $ 1372.6~\!(43.4) $ & $ 697.5~\!(2.2) $ & $ 0.0~\!(0.0) $ & $ 861.5~\!(27.3) $ \\
    & \!\!\textit{Q-Learning}\!\! & $ 1.0~\!(0.0) $ & $ 143.3~\!(4.4) $ & $ 459.2~\!(14.1) $ & $ 1318.1~\!(40.3) $ & $ 981.5~\!(30.0) $ & $ 0.0~\!(0.0) $ & $ 364.1~\!(11.1) $ \\
  
      \midrule
     & \textit{GDQ} & $ \!1025.1~\!(31.7)\! $ & $ 984.9~\!(30.5) $ & $ 621.8~\!(19.3) $ & $ \mathbf{ 13.6~\!(0.4) }$ & $ 7.5~\!(0.2) $ & $ 575.4~\!(17.8) $ & $ 2.5~\!(0.1) $ \\
    Task C & \!\textit{DARLING}\! & $ \!1343.9~\!(40.9)\! $ & $ 832.4~\!(25.4) $ & $ 582.1~\!(17.7) $ & $ 22.2~\!(0.7) $ & $ \mathbf{ 6.9~\!(0.2) }$ & $ 491.6~\!(15.0) $ & $ 3.2~\!(0.1) $ \\
    4,5,7 & \textit{Dyna-Q}  & $ \!1943.0~\!(43.5)\! $ & $ 854.9~\!(19.1) $ & $ 861.9~\!(19.3) $ & $ 203.5~\!(4.6) $ & $ 101.1~\!(2.3) $ & $ 495.3~\!(11.1) $ & $ 6.0~\!(0.1) $ \\
    & \!\!\textit{Q-Learning}\!\! & $ \!3146.4~\!(59.7)\! $ & $ 472.0~\!(9.0) $ & $ \!1143.0~\!(21.7)\! $ & $ 378.4~\!(7.2) $ & $ 125.7~\!(2.4) $ & $ 0.5~\!(0.0) $ & $ \mathbf{ 0.6~\!(0.0) }$ \\
  
    \midrule
     & \textit{GDQ} & $ \!1031.2~\!(31.8)\! $ & $ \mathbf{ 37.2~\!(1.1) }$ & $ 637.4~\!(19.6) $ &  $ 943.4~\!(29.1) $  & $ \mathbf{ 6.7~\!(0.2) }$ & $ 20.5~\!(0.6) $ & $ 570.0~\!(17.6) $ \\
    Task D & \!\textit{DARLING}\! & $ \!1503.7~\!(43.5)\! $ & $ 282.1~\!(8.2) $ & $ 578.9~\!(16.7) $ & $ 553.5~\!(16.0) $ & $ 12.3~\!(0.4) $ & $ 133.9~\!(3.9) $ & $ 394.3~\!(11.4) $ \\
    2, 5 & \textit{Dyna-Q} & $ \!1981.8~\!(44.0)\! $ & $ 395.0~\!(8.8) $ & $ 882.4~\!(19.6) $ & $ 638.3~\!(14.2) $ & $ 169.2~\!(3.8) $ & $ 98.4~\!(2.2) $ & $ 344.0~\!(7.6) $ \\
    & \!\!\textit{Q-Learning}\!\! & $ \!3152.3~\!(59.8)\! $ & $ 479.8~\!(9.1) $ & $ \!1140.7~\!(21.6)\! $ & $ 374.7~\!(7.1) $ & $ 127.4~\!(2.4) $ & $ 0.6~\!(0.0) $ & $ 0.5~\!(0.0) $ \\
\bottomrule
  \end{tabular}
  \end{small}
\end{table*}

\paragraph{Switching Task}

To evaluate Hypothesis III, we studied four scenarios where the robot's goal is changed after 2500 episodes of training. 
This experiment was repeated 10 times for computing the averages and standard errors as presented in Figure~\ref{fig:exp_switch}. 
The two subfigures show the results collected from two cases of task changes. 
We can see that GDQ can adapt to the task change and learn an optimal policy much faster than the baselines, leveraging the learned task-independent transition function.

\subsection{Real Robot Experiment}
We have conducted experiments using a Segway-based mobile robot platform (Figure~\ref{fig:map} on the right).
In the real world, the robot has to ask people to help open doors, where the action cost and success rate are noisy and out of our control. 
We forbade the robot from entering Area-5 in real-world experiments, because it is a long corridor, and navigating through that area takes a very long time.  
The following parameters are used in the real-robot experiment: 
$ R_{max} = 1000 $, $\alpha = 0.5$, $\gamma=0.95$, and $\epsilon=0.1$.
The robot's task is $M_X$ = (P5, P3), referred to as Task-$X$. 

Different from simulation experiments, we use time to measure the cost of navigation and door-opening actions (instead of a predefined fixed value). 
A maximum of $10$ steps is allowed, i.e., if the robot cannot complete Task-$X$ in 10 steps, the corresponding trial will be deemed unsuccessful. 
We have conducted a total of $30$ trails using the real robot. 

Each trial took up to 30 minutes to complete. 
The Segway-based robot runs out of battery in about five hours, and the experiments were conducted on three consecutive days (5 hours a day, and 15 hours in total). 
Due to the long time required for each trial (especially in the early learning phase), we only compared GDQ with one baseline. %of Dyna-Q. 
% For each approach (GDQ and Dyna-Q), we conducted experiments of 15 hours in total. 

\begin{figure}[t]
     \centering
        \includegraphics[width=0.7\columnwidth]{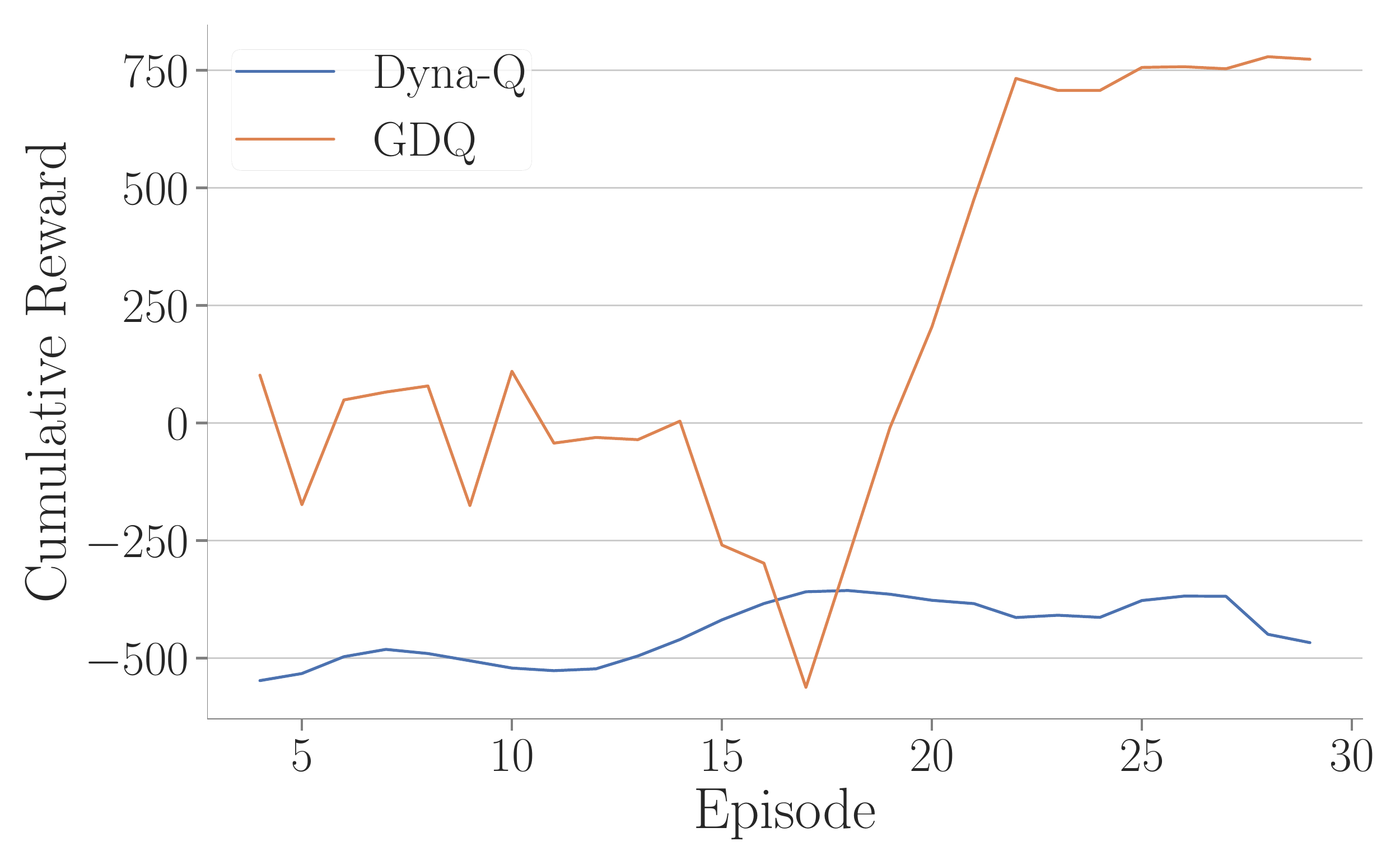}
        \caption{Task $M_X=(P5,P3)$ on a real robot.
        GDQ enabled the robot to find the optimal path in 22 trials, while Dyna-Q could not find a meaningful solution in 30 trials. }
     \label{fig:real}   
\end{figure}

Figure~\ref{fig:real} reports the results collected from the real-robot experiment. 
Looking at the very left of the two curves, the ``jump start'' of GDQ shows that Algorithm~\ref{alg:Algorithm1} (\textsc{OptInit}) helped the robot successfully avoid the ``random'' exploration behaviors in the early phase. 
Once the robot started interacting with the real world, we can see the cumulative reward of GDQ is consistently higher than Dyna-Q, except for only the $17th$ episode. 
After that, GDQ soon found the optimal solution. 
In comparison, Dyna-Q could not find a meaningful solution within a total of $30$ episodes.

\begin{figure}[t]
     \centering
        \includegraphics[width=0.8\columnwidth]{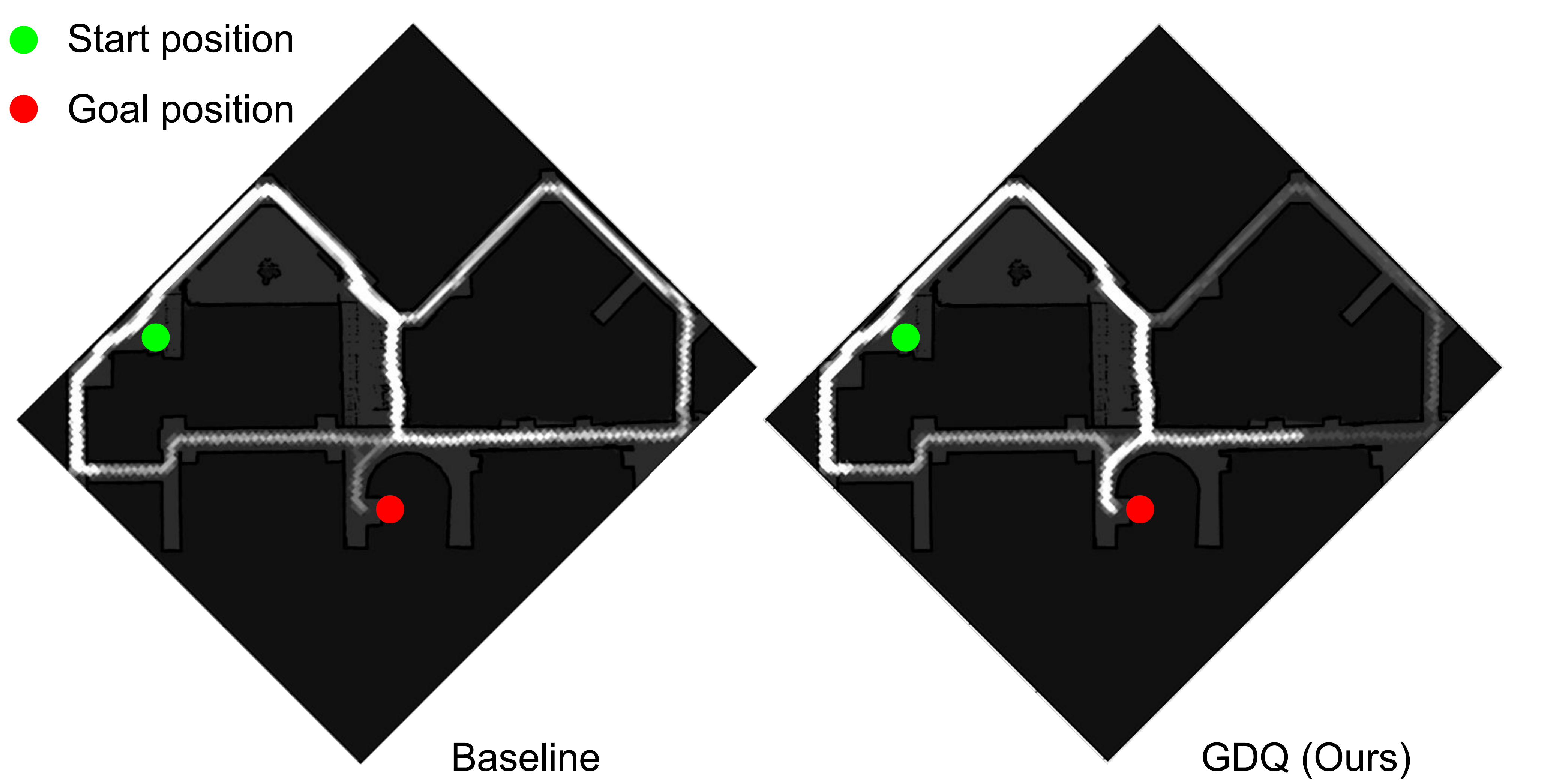}
        \caption{Heatmaps of a subarea of our office domain for visualizing where the robot visited using the Dyna-Q baseline (Left) and GDQ (Right). }
    \label{fig:heat}
\end{figure}

Figure~\ref{fig:heat} visualizes the frequency of our robot visiting different locations, where a light gray color represents a lower frequency of visits. 
We see that GDQ enabled the robot to focus more on the left side of the subarea, whereas, using the baseline approach, the robot traversed the right subarea (irrelevant) more.
We have generated a video for the demonstration of GDQ's performance on a real robot.\footnote{\verb https://youtu.be/X_Lc-8CD8No }

\section{Related Work}
\label{sec:related}
Automated planning is a branch of automated reasoning research that aims to compute a sequence of actions to accomplish complex tasks.
Automated planning methods frequently assume that the agent always gets the desired action outcomes, and unexpected outcomes are handled by plan monitoring and replanning. 
In comparison, RL methods assume non-deterministic action outcomes, and agents learn from interaction experience.  
We briefly summarize existing algorithms that leverage automated planning
~\cite{ghallab2016automated}
to improve the performance of reinforcement learning methods
~\cite{sutton2018reinforcement}
. 

\paragraph{Knowledge-based RL:}

Researchers have developed algorithms to integrate model-free RL and automated planning to avoid taking unreasonable actions in exploration. 
Algorithm DARLING is perhaps the earliest work that leverages action preconditions and effects from human knowledge for RL agents to avoid visiting risky or useless state, and has been applied to mobile robot navigation, and grid world domains~\cite{DBLP:journals/ai/LeonettiIS16}. 
Researchers have integrated automated planning and Q-learning focusing on non-stationary domains under uncertainties~\cite{ferreira2017answer}. 
Those algorithms exploited the flexibility of RL approaches and the accuracy of the declarative knowledge from humans. 
Other algorithms use action knowledge to improve model-free RL agents' performance in exploratory behaviors~\cite{ferreira2019solving,zhang2019faster}. 
% For instance, the work of~\citeauthor{zhang2019faster} discouraged a flappy bird from taking action ``flap'' when the bird is close to obstacles above it. 
In these works, researchers exploited the pre-designed models for constraining the state or action spaces. 
In comparison, GDQ (ours) equips the RL agent with the capability of simulating optimistic interaction experience using action knowledge for model learning and policy learning purposes. 

Researchers have developed algorithms to use subgoals to guide RL agents. 
Those subgoals can be learned and represented using non-monotonic logics~\cite{furelos2020induction}, or action languages~\cite{efthymiadis2013using}. 
The main difference from the above-mentioned methods is that GDQ uses model-based RL, whereas they used model-free RL methods that are task-oriented. 
Our service robotics domain includes potentially many service requests, rendering task-independent methods more suitable. 
Recently, \citeauthor{zhang2020survey} researches on leveraging knowledge to improve RL agents' learning performance.
% A recent paper surveyed research on leveraging knowledge to improve RL agents' learning performance~\cite{zhang2020survey}. 

% Recently, researchers have used policy gradient RL methods to guide classical planning~\cite{gomoluch2019learning,junyent2019deep}. 
% For instance, the work by \citeauthor{junyent2019deep} (\citeyear{junyent2019deep}) used policy gradient RL to select the best forward search methods for plan generation. 
% Their solutions are in the form of an action sequence from the classical planner, which do not consider the uncertainty in action outcomes. 
% GDQ 
% % The difference between these works and GDQ is that they are using RL to guide the planner, assuming deterministic action execution, while we are using 
% uses action knowledge to guide the behaviors of RL agents to account for the domain uncertainty. 

\paragraph{Hierarchical RL and Automated Planning:}
Planning methods have been used to guide the higher level of hierarchical RL methods~\cite{icarte2018using,yang2018peorl,lyu2019sdrl,jiang2019task,illanes2020symbolic,gordon2019should}. 
In those methods, the agents use an action language to compute plans to decompose a complex task into a sequence of subtasks, and each subtask is then implemented by a reinforcement learner. 
For instance, the work of \citeauthor{jiang2019task} (\citeyear{jiang2019task}) showed that the introduction of a few milestone positions at the task level can improve mobile robots' performance in indoor navigation tasks. 
The work of \citeauthor{icarte2018using} (\citeyear{icarte2018using}) built reward machines using temporal knowledge from domain experts to guide RL agents' learning behaviors, and also the reward machine can be learned from trial-and-error experiences~\cite{toro2019learning}. 

The domain knowledge used in those works significantly improved the learning efficiency of RL agents. 
% However, designing the hierarchy is frequently difficult, and many of the hierarchical methods trade optimality for efficiency (learning efficiency in this case). 
However, designing the hierarchy is frequently difficult, and many of the hierarchical methods trade optimality for learning efficiency. 
In comparison, GDQ uses action knowledge that is either publicly available (in our case) or can be easily encoded. 
Also, the optimistic experience of GDQ generated using action knowledge does not introduce any hard constraints, so GDQ inherits the optimality guarantee from RL algorithms. 

\paragraph{Logical Probabilistic Paradigms:}
There is the fundamental ``logic-probability'' gap between model-based RL and automated planning, where model-based RL relies on probabilistic transition systems, and traditionally automated planning does not model quantitative uncertainty. 
Aiming at bridging this gap, automated planning researchers have used logical-probabilistic paradigms to represent action knowledge, so as to directly reason about probabilistic transitions for model-based RL. 
% Researchers have studied model-based RL to improve the performance of automated planning by learning world models and refining action models~\cite{ijcai2019-443}. 
For instance, \citeauthor{ijcai2019-443}~(\citeyear{ijcai2019-443}) recently developed an algorithm that generates and updates logical-probabilistic action models of automated planning using model-based RL.
They used Probabilistic PDDL~\cite{younes2004ppddl1} for action modeling. 
Alternatively, researchers have developed new knowledge representation paradigms to help agents simultaneously reason with human knowledge and learn the model through interaction with the environment~\cite{wang2019bridging,lu2020learning,sridharan2019reba,veiga2019hierarchical,sanner2010symbolic}. 
% These approaches can improve the accuracy of planning by applying model-based RL to automated planning.
The above-mentioned methods require the human developer to manually encode logical-probabilistic knowledge, which requires significant professional skills and might soon become infeasible in large domains. 
In comparison, GDQ requires the minimum amount of action knowledge (widely available in our case), such as ``\emph{After going through a door, a robot will be on the other side of it}~\cite{yang2014planning,jiang2019comparison}, rendering GDQ more applicable to real-world domains.

% RL algorithms incorporating automated planning have also been conducted in complex domains. 
% SDRL has introduced symbolic planning into deep RL and enabled its the agent to deal with a high-dimensional state-space~\cite{DBLP:journals/corr/abs-1811-00090}. 
% Researchers also developed frameworks that rely on human guidance to solve complex deep RL tasks~\cite{zhang2019leveraging}. 
% However, these approaches are goal-oriented and need a significant amount of sampling effort. 
% In a continuous domain, \citeauthor{faust2018prm} have developed PRM-RL aiming at overcoming the computational limitation in the long-range navigation tasks using a real robot. 
% The planning of PRM-RL is at motion level and not able to incorporate human knowledge.
% However, in this paper, we focus on a navigation task and use a planner at task level.

% Our approach takes complementary features both of model-based RL and automated planning, aiming at improving sample-efficiency in a real robot domain.
% Our significant contribution apart from these approaches is how efficiently the agent can explore states of the domain, avoiding less-relevant states once by reasoning with contextual knowledge while trial-and-error experiences.

% \commenty{to here--------------------------------------------------------- }
\section{Conclusions}
In this paper, we develop Guided Dyna-Q (GDQ) for bridging the gap between model-based RL, and automated planning. 
The goal is to help the agent (robot) avoid exploring less-relevant states toward speeding up the learning process. 
GDQ has been demonstrated and evaluated both in simulation and using a real robot conducting navigation tasks in an indoor office environment. 
From the experimental results, we see that, using the widely available action knowledge, GDQ performed significantly better than competitive baseline methods from the literature, demonstrating the best performance in learning efficiency. 
% This work presents a novel algorithm for integrating the two paradigms of model-based RL and automated planning. 
% In the future, we evaluate how the action knowledge's granularity and accuracy affect the system performance. 
% For instance, we are interested in evaluating if GDQ is able to recover from outdated knowledge, and its sensitivity. 
% Besides, we would like to study how to minimize human effort in knowledge encoding while maximizing its benefit to reinforcement learning, focusing on the practical trade-off between the effort required to encode human knowledge and system performance.

\section*{Acknowledgment}
This work has taken place in the Autonomous Intelligent Robotics (AIR) Group at SUNY Binghamton. AIR research is supported in part by grants from the National Science Foundation (NRI-1925044), Ford Motor Company (URP Awards 2019 and 2020), OPPO (Faculty Research Award 2020), and SUNY Research Foundation. 

{
\bibliography{references}
}
% \bigskip

\end{document}